\pdfoutput=1
\documentclass[11pt]{article}
\usepackage[final]{naacl2021}
\usepackage{times}
\usepackage{latexsym}
\usepackage[T1]{fontenc}
\usepackage[utf8]{inputenc}

\usepackage{pgfplots}
\pgfplotsset{compat=1.14} 
\usepackage{graphicx}
\usepackage{xspace}
\usepackage{url}
\usepackage{pifont}
\usepackage{amsmath}
\usepackage{textcomp}
\usepackage{multirow}
\usepackage{graphicx}
\usepackage{microtype}
\usepackage{enumitem}
\usepackage{array}
\newcolumntype{L}[1]{>{\raggedright\let\newline\\\arraybackslash\hspace{0pt}}m{#1}}
\newcolumntype{C}[1]{>{\centering\let\newline\\\arraybackslash\hspace{0pt}}m{#1}}
\newcolumntype{R}[1]{>{\raggedleft\let\newline\\\arraybackslash\hspace{0pt}}m{#1}}

\newcommand{\rnn}{\textsc{rnn}\xspace}

\newcommand{\nlp}{\textsc{nlp}\xspace}
\newcommand{\echr}{\textsc{echr}\xspace}
\newcommand{\ecthr}{\textsc{ec}t\textsc{hr}\xspace}
\newcommand{\gan}{\textsc{gan}\xspace}
\newcommand{\bert}{\textsc{bert}\xspace}
\newcommand{\cls}{\textsc{[cls]}\xspace}
\newcommand\selu{\textsc{selu}\xspace}
\newcommand{\hierbertmax}{\textsc{hierbert-all}\xspace}
\newcommand{\hierbert}{\textsc{hierbert}\xspace}
\newcommand{\hierbertsa}{\textsc{hierbert-sa}\xspace}
\newcommand{\hierbertha}{\textsc{hierbert-ha}\xspace}
\newcommand{\huggingface}{
    \includegraphics[scale=0.2]{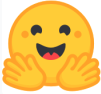}
}

\title{Paragraph-level Rationale Extraction through Regularization: \\ A case study on European Court of Human Rights Cases}

\author{Ilias Chalkidis$^{\;\dagger\;\ddagger}$\qquad Manos Fergadiotis$^{\;\dagger\;\ddagger}$ \qquad Dimitrios Tsarapatsanis$^{\;\star}$\\
\textbf{Nikolaos Aletras$^{\;\diamond}$} \qquad \textbf{Ion Androutsopoulos$^{\;\ddagger\;\dagger}$} \qquad \textbf{Prodromos Malakasiotis$^{\;\dagger\;\ddagger}$} \\
$^{\dagger\;}$EY AI Centre of Excellence in Document Intelligence, NCSR ``Demokritos'' \\ $^{\ddagger\;}$Department of Informatics, Athens University of Economics and Business \\
$^\diamond$ Computer Science Department, University of Sheffield \\
$^\star$ Law School, University of York}

\date{}

\begin{document}
\maketitle
\begin{abstract}
\emph{Interpretability} or \emph{explainability} is an emerging research field in \nlp. From a \emph{user-centric} point of view, the goal is to build models that provide proper justification for their decisions, similar to those of humans, by requiring the models to satisfy additional constraints. To this end, we introduce a new application on legal text where, contrary to mainstream literature targeting word-level rationales, we conceive rationales as selected \emph{paragraphs} in multi-paragraph structured court cases. We also release a new dataset comprising European Court of Human Rights cases, including annotations for paragraph-level rationales. We use this dataset to study the effect of already proposed rationale constraints, i.e., \emph{sparsity}, \emph{continuity}, and \emph{comprehensiveness}, formulated as regularizers. Our findings indicate that some of these constraints are not beneficial in paragraph-level rationale extraction, while others need re-formulation to better handle the multi-label nature of the task we consider. We also introduce a new constraint, \emph{singularity}, which further improves the quality of rationales, even compared with noisy rationale supervision. Experimental results indicate that the newly introduced task is very challenging and there is a large scope for further research.  
{\let\thefootnote\relax\footnotetext{Correspondence to: {\tt ihalk.aueb.gr}}}
\end{abstract}

\begin{figure*}
    \centering
    \includegraphics[width=0.95\textwidth]{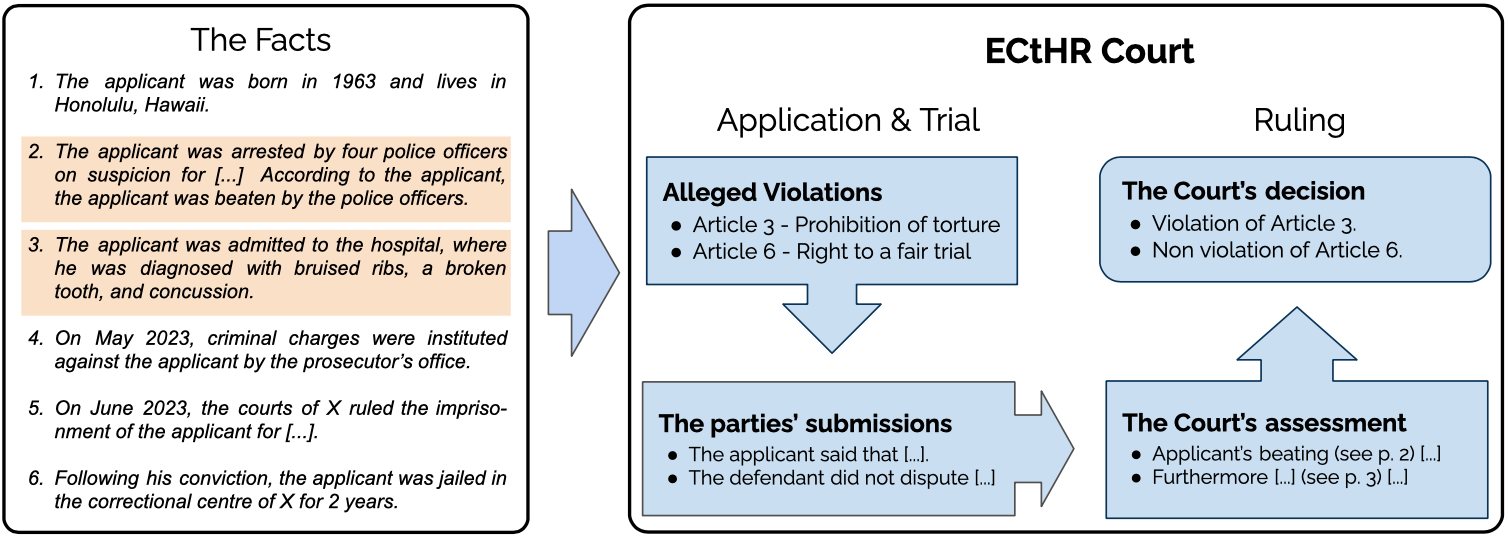}
    \caption{A depiction of the \ecthr process: The applicant(s) request a hearing from \ecthr regarding specific accusations (alleged violations of \echr articles) against the defendant state(s), based on facts. The Court (judges) assesses the facts and the rest of the parties' submissions, and rules on the violation or not of the allegedly violated \echr articles. Here, prominent facts referred in the court's assessment are highlighted.}
    \label{fig:echr_process}
    \vspace{-4mm}
\end{figure*}

\section{Introduction} \label{sec:intro}

Model \emph{interpretability} (or \emph{explainability}) is an emerging field of research in \nlp  \cite{lipton2018,jacovi2020}. From a \emph{model-centric} point of view, the main focus is to demystify a model's inner workings, for example targeting  self-attention mechanisms \cite{Jain2019,wiegreffe2019}, and more recently Transformer-based language models \cite{clark-etal-2019-bert, kovaleva-etal-2019-revealing, rogers2020primer}. From a \emph{user-centric} point of view, the main focus is to build models that learn to provide proper justification for their decisions, similar to those of humans, \cite{zaidan2007, lei2016, chang2019, yu2019} by requiring the models to satisfy additional constraints. 

Here we follow a \emph{user-centric} approach to \emph{rationale extraction}, where the model learns to select a subset of the input that justifies its decision. To this end, we introduce a new application on legal text where, contrary to mainstream literature targeting word-level rationales, we conceive rationales as automatically selected  paragraphs in multi-paragraph structured court cases. While previous related work targets mostly binary text classification tasks \cite{deyoung2020}, our task is a highly skewed multi-label text classification task.  Given a set of paragraphs that refer to the facts of each case (henceforth \emph{facts}) in judgments of the European Court of Human Rights (\ecthr), the model aims to predict the \emph{allegedly violated} articles of the European Convention of Human Rights (\echr). We adopt a \emph{rationalization by construction} methodology \cite{lei2016, chang2019, yu2019}, where the model is regularized to satisfy additional constraints that reward the model, if its decisions are based on concise rationales it selects, as opposed to inferring explanations from the model's decisions in a post-hoc manner \cite{RibeiroSG16,alvarez2017,Murdoch2018}.

\emph{Legal judgment prediction} has been studied in the past for cases ruled by the European Court of Human Rights \cite{Aletras2016, Medvedeva2018, Chalkidis2019} and for Chinese criminal court cases \cite{Luo2017, Hu2018, Zhong2018}, but there is no precedent of work investigating the justification of the models' decisions. Similarly to other domains (e.g., financial, biomedical), explainability is a key feature in the legal domain, which may potentially improve the trustworthiness of systems that abide by the principle of the \emph{right to explanation} \cite{goodman2017}. We investigate the explainability of the decisions of state-of-the-art models, comparing the paragraphs they select to those of legal professionals, both litigants and lawyers, in \emph{alleged violation prediction}. In the latter task, introduced in this paper, the goal is to predict the accusations (allegations) made by the applicants. The accusations can be usually predicted given only the facts of each case. By contrast, in the previously studied legal judgment prediction task, the goal is to predict the court's decision; this is much more difficult and vastly relies on case law (precedent cases). 

Although the new task (alleged violation prediction) is simpler than legal judgment prediction, models that address it (and their rationales) can still be useful in the judicial process (Fig.~\ref{fig:echr_process}). For example, they can help applicants (plaintiffs) identify alleged violations that are supported by the facts of a case. They can help judges identify more quickly facts that support the alleged violations, contributing towards more informed judicial decision making \cite{zhong2020}. They can also help legal experts identify previous cases related to particular allegations, helping analyze case law \cite{Katz2012}.  
Our contributions are the following:

\begin{itemize}[leftmargin=8pt]
    \itemsep0em
    \item We introduce \emph{rationale extraction} for \emph{alleged violation prediction} in \ecthr cases, a more tractable task compared to legal judgment prediction. This is a \emph{multi-label} classification task that requires \emph{paragraph-level} rationales, unlike previous work on word-level rationales for binary classification.
    \item We study the effect of previously proposed rationale constraints, i.e., \emph{sparsity}, \emph{continuity} \cite{lei2016}, and \emph{comprehensiveness} \cite{yu2019}, formulated as regularizers. We show that \emph{continuity} is not beneficial and requisite in paragraph-level rationale-extraction, while \emph{comprehensiveness} needs to be re-formulated for the multi-label nature of the task we consider. We also introduce a new constraint, \emph{singularity}, which further improves the rationales, even compared with silver (noisy) rationale supervision.
    \item We release a new dataset for alleged article violation prediction, comprising 11k \ecthr cases in English, with silver rationales obtained from references in court decisions, and gold rationales provided by \echr{-experienced} lawyers.\footnote{Our dataset is publicly available at \url{https://huggingface.co/datasets/ecthr_cases}, see usage example in Appendix~\ref{sec:appendix_e}.}
\end{itemize}

\noindent To the best of our knowledge, this is also the first work on rationale extraction that fine-tunes end-to-end pre-trained Transformer-based models.\footnote{Others fine-tuned such models only partially \cite{jain2020}, i.e., top two layers, or not at all \cite{deyoung2020}.}

\section{Related Work} \label{sec:relatedWork}

\noindent\textbf{Legal judgment prediction:} Initial work on legal judgment prediction in English used linear models with features based on bags of words and topics, applying them to \ecthr cases \citep{Aletras2016,Medvedeva2018}. More recently, we experimented with neural methods \cite{Chalkidis2019} , showing that hierarchical \rnn{s} \citep{Yang2016}, and a hierarchical variation of \textsc{bert} \citep{devlin2019} that encodes paragraphs, outperform linear classifiers with bag-of-word representations.

In all previous work, legal judgment prediction is tackled in an over-simplified experimental setup where only textual information from the cases themselves is considered, ignoring many other important factors that judges consider, more importantly general legal argument and past case law. Also, \citet{Aletras2016}, \citet{Medvedeva2018}, \citet{Chalkidis2019} treat \ecthr judgment prediction as a binary classification task per case (any article violation or not), while the \ecthr actually considers and rules on the violation of individual articles of the European Convention of Human Rights (\echr). 

In previous work \cite{Chalkidis2019}, we also attempted to predict which \emph{particular} articles were violated, assuming, however, that the Court considers all the \echr articles in each case, which is not true. In reality, the Court considers only alleged violations of particular articles, argued by applicants. Establishing which articles are allegedly violated is an important preliminary task when preparing an \ecthr application. Instead of oversimplifying the overall judgment prediction task, we focus on the preliminary task and use it as a test-bed for generating paragraph-level rationales in a multi-label text classification task for the first time.

Legal judgment prediction has also been studied in Chinese criminal cases \citep{Luo2017, Hu2018,Zhong2018}. Similarly to the literature on legal judgment prediction for \ecthr cases, the aforementioned approaches ignore the crucial aspect of justifying the models' predictions. 

Given the gravity that legal outcomes have for individuals, explainability is essential to increase the trust of both legal professionals and laypersons on system decisions and promote the use of supportive tools \cite{barfield2020}. To the best of our knowledge, our work is the first step towards this direction for the legal domain, but is also applicable in other domains (e.g., biomedical), where justifications of automated decisions are essential.
% components of both legitimacy and acceptability.
\vspace{2mm}

\noindent\textbf{Rationale extraction by construction:} Contrary to earlier work that required supervision in the form of human-annotated rationales \citep{zaidan2007,zhang2016}, \citet{lei2016} introduced a \emph{self-supervised} methodology to extract rationales (that supported aspect-based sentiment analysis predictions), i.e., gold rationale annotations were used only for evaluation. Furthermore, models were designed to produce rationales \emph{by construction}, contrary to work studying saliency maps (generated by a model without explainability constraints) using gradients or perturbations at inference time \cite{RibeiroSG16,alvarez2017,Murdoch2018}. \citet{lei2016} aimed to produce short coherent rationales that could replace the original full texts,  maintaining the model's predictive performance. The rationales were extracted by generating binary masks indicating which words should be selected; and two additional loss regularizers were introduced, which penalize long rationales and sparse masks (that would select non-consecutive words). 

\citet{yu2019} proposed another constraint to ensure that the rationales would contain all the relevant information. They formulated this constraint through a \emph{minimax} game, where two players, one using the predicted binary mask and another using the complement of this mask, aim to correctly classify the text. If the first player fails to outperform the second, the model is penalized.
\citet{chang2019} use a Generative Adversarial Network (\gan) \cite{goodfellow2014}, where a generator producing factual rationales competes with a generator producing counterfactual rationales to trick a discriminator. The \gan was not designed to perform classification. Given a text and a label it produces a rationale supporting (or not) the label. 

\citet{jain2020} decoupled the model's predictor from the rationale extractor to produce inherently faithful explanations, ensuring that the predictor  considers only the rationales and not other parts of the text. Faithfulness refers to how accurately an explanation reflects the true reasoning of a model \cite{lipton2018, jacovi2020}.

All the aforementioned work conceives rationales as selections of words, targeting binary classification tasks even when this is inappropriate. For instance, \citet{deyoung2020} and \citet{jain2020} over-simplified the task of the multi-passage reading comprehension (MultiRC) dataset \citep{MultiRC2018} turning it into a binary classification task with word-level rationales, while sentence-level rationales seem more suitable. 

\noindent\textbf{Responsible AI:} Our work complies with the \ecthr data policy.
% \footnote{\url{https://www.echr.coe.int/Pages/home.aspx?p=privacy}}  
By no means do we aim to build a `robot' lawyer or judge, and we acknowledge the possible harmful impact \cite{angwin2016,dressel2018} of irresponsible deployment. Instead, we aim to support fair and explainable \textsc{ai}-assisted judicial decision making and empirical legal studies.
% , as in the use cases suggested in Section~\ref{sec:intro}.
We consider our work as part of ongoing critical research on responsible \textsc{ai} \cite{facct2021} that aims to provide explainable and fair  systems to support human experts.

\begin{table}[h]
    \footnotesize{
    \centering
    \resizebox{\columnwidth}{!}{
    \begin{tabular}{l|c|c|c}
         &  Cases & Sparsity & \#Allegations \\ \hline
         Train & 9K & 24\% & 1.8 \\
         Development & 1K & 30\% & 1.7 \\
         Test & 1K & 31\% & 1.7 \\
    \end{tabular}
    }
        \vspace*{-2mm}
    \caption{Statistics of the new \ecthr dataset. `Sparsity' is the average percentage of paragraphs included in the silver rationales. `\#Allegations' is the average number of allegedly violated articles.}
        \vspace*{-5mm}
    \label{tab:dataset}
    }
\end{table}

\section{The New ECtHR Dataset}
\label{sec:dataset}

The court (\ecthr) hears allegations regarding breaches in human rights provisions of the European Convention of Human Rights (\echr) by European states (Fig.~\ref{fig:echr_process}).\footnote{The Convention is available at \url{https://www.echr.coe.int/Documents/Convention_ENG.pdf}.} 
The court rules on a subset of all \echr articles, which are predefined (alleged) by the applicants (\emph{plaintiffs}). Our dataset comprises 11k \ecthr cases and can be viewed as an enriched version of the \ecthr dataset of \citet{Chalkidis2019}, which did not provide ground truth for alleged article violations (articles discussed) and rationales. The new dataset includes the following:\vspace{1.5mm}

\noindent\textbf{Facts:} Each judgment includes a list of paragraphs that represent the facts of the case, i.e., they describe the main events that are relevant to the case, in numbered paragraphs. We hereafter call these paragraphs \emph{facts} for simplicity. Note that the facts are presented in chronological order. Not all facts have the same impact or hold crucial information with respect to alleged article violations and the court's assessment; i.e., facts may refer to information that is trivial or otherwise irrelevant to the legally crucial allegations against \emph{defendant} states.\vspace{1.5mm}

\noindent\textbf{Allegedly violated articles:} Judges rule on specific accusations (allegations) made by the applicants \cite{oboyle2018}. In \ecthr cases, the judges discuss and rule on the violation, or not, of specific articles of the Convention. The articles to be discussed (and ruled on) are put forward (as alleged article violations) by the applicants and are included in the dataset as ground truth; we identify 40 violable articles in total.\footnote{The rest of the articles are procedural, i.e., the number of judges, criteria for office, election of judges, etc.} In our experiments, however, the models are not aware of the allegations. They predict the Convention articles that will be discussed (the allegations) based on the case's facts, and they also produce rationales for their predictions. Models of this kind could be used by potential applicants to help them formulate future allegations (articles they could claim to have been violated), as already noted, but here we mainly use the task as a test-bed for rationale extraction.\vspace{1.5mm}

\noindent\textbf{Violated articles:} The court decides which allegedly violated articles have indeed been violated. These decisions are also included in our dataset and could be used for full legal judgment prediction experiments \cite{Chalkidis2019}. However, they are not used in the experiments of this work.\vspace{1.5mm}

\noindent\textbf{Silver allegation rationales:} Each decision of the \ecthr includes references to facts of the case (e.g., ``\emph{See paragraphs 2 and 4}.'') and case law (e.g., ``\emph{See Draci vs.\ Russia (2010)}.''). We identified references to each case's facts and retrieved the corresponding paragraphs using regular expressions. These are included in the dataset as silver allegation rationales, on the grounds that the judges refer to these paragraphs when ruling on the allegations.\vspace{1.5mm}

\noindent\textbf{Gold allegation rationales:} A legal expert with experience in \ecthr cases annotated a subset of 50 test cases to identify the relevant facts (paragraphs) of the case that support the allegations (alleged article violations). In other words, each identified fact justifies (hints) one or more alleged violations.\footnote{For details on the annotation process and examples of annotated \ecthr cases, see Appendices~\ref{sec:appendix_c}, \ref{sec:appendix_f}.}\vspace{1.5mm}

\noindent\textbf{Task definition:} In this work, we investigate \emph{alleged violation prediction}, a multi-label text classification task where, given the facts of a \ecthr case, a model predicts which of the 40 violable \echr articles were allegedly violated according to the applicant(s).\footnotemark[4] The model also needs to identify the facts that most prominently support its decision.

\section{Methods}

We first describe a baseline model that we use as our starting point. It adopts the framework proposed by \citet{lei2016}, which generates rationales by construction: a \emph{text encoder} sub-network reads the text; a \emph{rationale extraction} sub-network produces a binary mask indicating the most important words of the text; and a \emph{prediction} sub-network classifies a hard-masked version of the text. We then discuss additional constraints that have been proposed to improve word-level rationales, which can be added to the baseline as regularizers. We argue that one of them is not beneficial for paragraph-level rationales. We also consider variants of previous constraints that better suit multi-label classification tasks and introduce a new one.

\subsection{Baseline Model}

\begin{figure}[h]
\begin{center}
\includegraphics[width=\columnwidth]{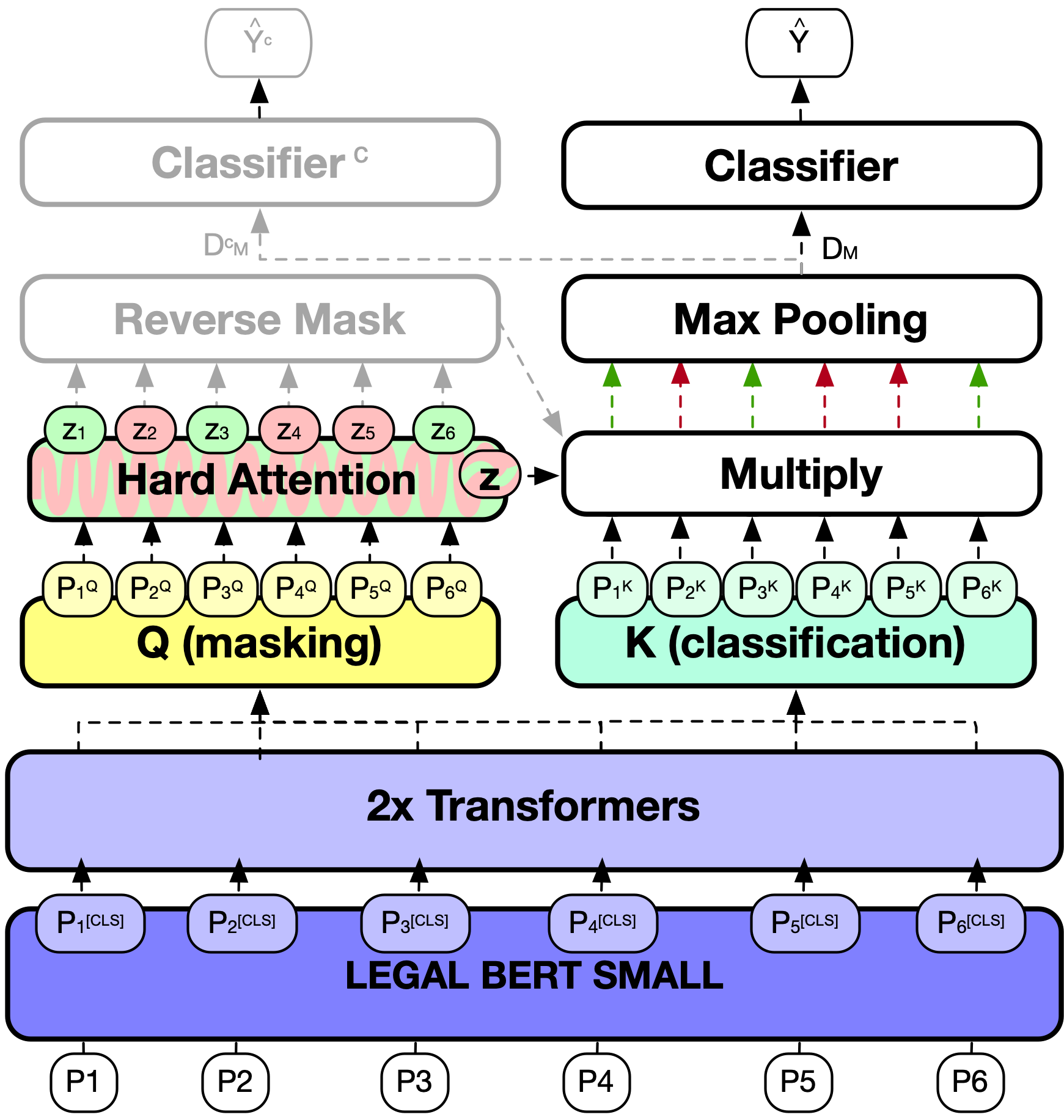}
\end{center}
\vspace*{-3mm}
\caption{Illustration of \hierbertha. The shaded parts operate only when $Lg$ or $Lr$ are used.}
\vspace*{-5mm}
\label{fig:model}
\end{figure}

Our baseline is a hierarchical variation of \bert \cite{devlin2019} with hard attention, dubbed \hierbertha.\footnote{In previous work, we proposed a hierarchical variation of \bert with self-attention \cite{Chalkidis2019}. In parallel work, \citet{yang2020} proposed a similar Transformer-based Hierarchical Encoder (SMITH) for long document matching.} Each case (document) $D$ is viewed as a list of facts (paragraphs) $D=[P_1, \dots, P_N]$. Each paragraph is a list of tokens $P_i=[w_1, \dots, w_{L_i}]$. We first pass each paragraph independently through a shared \bert encoder (Fig.~\ref{fig:model}) to extract context-unaware paragraph representations $P^{\cls}_i$, using the \cls embedding of \bert. Then, a shallow encoder with two Transformer layers \cite{Vaswani2017} produces contextualized paragraph embeddings, which are in turn projected to two separate spaces by two different fully-connected layers, $K$ and $Q$, with \selu activations \cite{KlambauerUMH17}. $K$ produces the paragraph encoding $P^K_i$, to be used for classification;  and $Q$ produces the paragraph encoding $P^Q_i$, to be used for rationale extraction. The rationale extraction sub-network passes each $P^Q_i$ encoding independently through a fully-connected layer with a sigmoid activation to produce soft attention scores $a_i\in[0,1]$. The attention scores are then binarized using a 0.5 threshold, leading to hard attention scores $z_i$ ($z_i = 1$ iff $a_i > 0.5$). The hard-masked document representation $D_M$ is obtained by hard-masking paragraphs and max-pooling: 
\begin{align}
\nonumber
    D_M = \mathrm{maxpool}\!\left(
    % Z \odot P^K
    [z_1 \cdot P_1^K, \dots, 
    z_N \cdot P_N^K]
    \right)
\end{align}
$D_M$ is then fed to a dense layer with sigmoid activations, which produces a probability estimate per label, $\widehat{Y} = [ \hat{y_1}, \dots, \hat{y_{|A|}}]$, in our case per article of the Convention, where $|A|$ is the size of the label set. For comparison, we also experiment with a model that masks no facts, dubbed \hierbertmax.

The thresholding that produces the hard (binary) masks $z_i$ is not differentiable. To address this problem, \citet{lei2016} used reinforcement learning \cite{williams1992}, while \citet{bastings2019} proposed a differentiable mechanism relying on the re-parameterization trick \cite{louizos2017}. We follow a simpler trick, originally proposed by \citet{chang2019}, where during backpropagation the thresholding is detached from the computation graph, allowing the gradients to bypass the thresholding and reach directly the soft attentions $a_i$.

\subsection{Rationale Constraints as Regularizers}
\label{sec:rationales}

\noindent\textbf{Sparsity:} Modifying the word-level sparsity constraint of  \citet{lei2016} for our paragraph-level rationales, we also hypothesize that good rationales include a small number of facts (paragraphs) that sufficiently justify the allegations; the other facts are trivial or secondary. For instance, an introductory fact like ``\emph{The applicant was born in 1984 and lives in Switzerland.}'' does not support any allegation, while a fact like ``\emph{The applicant contended that he had been beaten by police officers immediately after his arrest and later during police questioning.}'' suggests a violation of Article 3 ``Prohibition of Torture''. Hence, we use a \emph{sparsity} loss to control the number of selected facts:
\begin{align}
L_{s} = \left|T - \frac{1}{N}\sum_{i=1}^{N} z_i  \right|
\label{eq:sparsity}
\end{align}
where $T$ is a predefined threshold specifying the desired percentage of selected facts per case. We can estimate $T$ from silver rationales (Table~\ref{tab:dataset}).\vspace{1.5mm}

\noindent\textbf{Continuity:}
In their work on word-level rationales, \citet{lei2016} also required  the selected words to be \emph{contiguous}, to obtain more coherent rationales. In other words, the transitions between selected ($z_i=1$) and not selected ($z_i = 0$) words in the hard mask should be minimized. This is achieved by adding the following \emph{continuity} loss:
\begin{align}
L_{c} = \frac{1}{N-1}\sum_{i=2}^{N}|z_i - z_{i-1}|
\end{align}
In paragraph-level rationale extraction, where entire paragraphs are masked, the continuity loss forces the model to select contiguous paragraphs. In \ecthr cases, however, the facts are self-contained and internally coherent paragraphs (or single sentences). Hence, we hypothesize that the \emph{continuity} loss is not beneficial in our case. Nonetheless, we empirically investigate its effect.\vspace{1.5mm}

\noindent\textbf{Comprehensiveness:} 
We also adapt the \emph{comprehensiveness} loss of 
\citet{yu2019}, which was introduced to force the hard mask $Z = [z_1, \dots, z_N]$ to (ideally) keep \emph{all} the words (in our case, paragraphs about facts) of the document $D$ that support the correct decision $Y$. In our task, $Y = [y_1, \dots, y_{|A|}]$ is a binary vector indicating the Convention articles the court discussed (gold allegations) in the case of $D$. Intuitively, the complement $Z^c$ of $Z$, i.e., the hard mask that selects the words (in our case, facts) that $Z$ does not select, should not select sufficient information to predict $Y$. Given $D$, let $D_M, D_M^c$ be the representations of $D$ obtained with $Z, Z^c$, respectively; let $\widehat{Y}, \widehat{Y}^c$ be the corresponding probability estimates; let $L_p, L_p^c$ be the classification loss, typically total binary cross-entropy, measuring how far $\widehat{Y}, \widehat{Y}^c$ are from $Y$. In its original form, the comprehensiveness loss requires $L_p^c$ to exceed $L_p$ by a margin $h$.
\begin{align}
    L_g = \max(L_p - L^c_p + h, 0)
    \label{eq:compr_entropy}
\end{align}
While this formulation may be adequate in binary classification tasks, in multi-label classification it is very hard to pre-select a reasonable margin, given that cross-entropy is unbounded, that the distribution of true labels (articles discussed) is highly skewed, and that some labels are easier to predict than others. To make the selection of $h$ more intuitive, we propose a reformulation of $L_g$ that operates on class probabilities rather than classification losses. The right-hand side of Eq.~\ref{eq:compr_entropy} becomes:
\begin{align}
    % L_g =
    \frac{1}{|A|}\!\sum_{i=1}^{|A|} y_i(\!\hat{y_i}^c\!-\!\hat{y_i}\!+\!h)\!+\!(1\!-\!y_i)(\hat{y_i}\!-\!\hat{y_i}^c\!\!+\!h)\label{eq:compr_prob}
\end{align}
The margin $h$ is now easier to grasp and tune. It encourages the same gap between the probabilities predicted with $Z$ and $Z^c$  across all labels (articles). 

We also experiment with a third variant of comprehensiveness, which does not compare the probabilities we obtain with $Z$ and $Z^c$, comparing instead the two latent document representations:
\begin{align}
    L_g = \left|\mathrm{cos}(D_M,  D^c_M)\right|
    \label{eq:compr_repr}
\end{align}
where $\mathrm{cos}$ denotes cosine similarity. This variant forces $D_M$ and $D^c_M$ to be as dissimilar as possible, without requiring a preset margin.
\vspace{2mm}

\noindent\textbf{Singularity:} A limitation of the comprehensiveness loss (any variant) is that it only requires the mask $Z$ to be better than its complement $Z^c$. This does not guarantee that $Z$ is better than \emph{every} other mask. Consider a case where the gold rationale identifies three articles and $Z$ selects only two of them. The model may produce better predictions with $Z$ than with $Z^c$, and $D_M$ may be very different than $D_M^c$ in Eq.~\ref{eq:compr_repr}, but $Z$ is still not the best mask. To address this limitation, we introduce the \emph{singularity} loss $L_r$, which requires $Z$ to be better than a mask $Z^r$, randomly generated per training instance and epoch, that selects as many facts as the sparsity threshold $T$ allows: 
\begin{flalign}
L_r &= \gamma \cdot L_g(Z, Z^r) \label{eq:sing} \\
\gamma &= 1 - \mathrm{cos}(Z^r, Z) \notag
\end{flalign}
Here $L_g(Z, Z^r)$ is any variant of $L_g$, but now using $Z^r$ instead of $Z^c$; and $\gamma$ regulates the effect of $L_g(Z, Z^r)$ by considering the cosine distance between $Z^r$ and $Z$. The more $Z$ and $Z^r$ overlap, the less we care if $Z$ performs better than $Z^r$. 

The total loss of our model is computed as follows. Again $L_p$ is the classification loss; $L_p^c, L_p^r$ are the classification losses when using $Z^c, Z^r$, respectively; and all $\lambda$s are tunable hyper-parameters.
\begin{align}
\label{eq:totalLoss}
L & = L_p + \lambda_s \cdot L_s + \lambda_c \cdot L_c\notag\\
  & + \lambda_g \, (L_g + L^c_p) + 
  \lambda_r \, (L_r + L^r_p)
\end{align}
We include $L^c_p$ in Eq.~\ref{eq:totalLoss}, because otherwise the network would have no incentive to make $D_M^c$ and $\widehat{Y}^c$ competitive in prediction; and similarly for $L^r_p$.\vspace{2mm} 

\noindent\textbf{Rationales supervision:} For completeness we also experimented with a variant that utilizes silver rationales for noisy rationale supervision \cite{zaidan2007}. In this case the total loss becomes:
\begin{align}
    L = L_p + \lambda_{ns} \cdot \textsc{mae}(Z, Z^s)
    \label{eq:noisy}
\end{align}
where \textsc{mae} is the mean absolute error between the predicted mask, $Z$, and the silver mask, $Z_s$, and $\lambda_{ns}$ weighs the effect of \textsc{mae} in the total loss.

\section{Experimental Setup}

For all methods, we conducted grid-search to tune the hyper-parameters $\lambda_*$. We used the Adam optimizer \citep{Kingma2015} across all experiments with a fixed learning rate of $2\mathrm{e}\text{-}5$.\footnote{In preliminary experiments, we tuned the baseline model on development data as a stand-alone classifier and found that the optimal learning rate was $2\mathrm{e}\text{-}5$, searching in the set \{$2\mathrm{e}\text{-}5$, $3\mathrm{e}\text{-}5$, $4\mathrm{e}\text{-}5$, $5\mathrm{e}\text{-}5$\}. The optimal drop-out rate was 0.} All methods rely on \textsc{legal-bert-small} \cite{chalkidis2020}, a variant of \bert \cite{devlin2019}, with 6 layers, 512 hidden units and 8 attention heads, pre-trained on legal corpora. Based on this model, we were able to use up to 50 paragraphs of 256 words each in a single 32\textsc{gb gpu}. In preliminary experiments, we found that the proposed model relying on a shared paragraph encoder, i.e., one that passes the same context-aware paragraph representations $P_i^{[CLS]}$ to both the $Q$ and $K$ sub-networks, as in Fig.~\ref{fig:model}, has comparable performance and better rationale quality, compared to a model with two independent paragraph encoders, as the one used in the literature \cite{lei2016, yu2019, jain2020}.\footnote{See Appendix~\ref{sec:appendix_b} for additional details and results.} 
For all experiments, we report the average and standard deviation across five runs.

We evaluate: (a) classification performance, (b) \emph{faithfulness} (Section~\ref{sec:relatedWork}), and (c) \emph{rationale quality}, while respecting a given sparsity threshold ($T$).\vspace{1.5mm}

\noindent\textbf{Classification performance:} Given the label skewness, we evaluate classification performance using \emph{micro-F1}, i.e., for each Convention article, we compute its F1, and micro-average over articles.\vspace{1.5mm}

\noindent\textbf{Faithfulness:} Recall that \emph{faithfulness} refers to how accurately an explanation reflects the true reasoning of a model. To measure faithfulness, we report \emph{sufficiency} and \emph{comprehensiveness} \cite{deyoung2020}. Sufficiency measures the difference between the predicted probabilities for the gold (positive) labels when the model is fed with the whole text ($\widehat{Y_+}^f$) 
and when the model is fed only with the predicted rationales ($\widehat{Y_+}$). Comprehensiveness (not to be confused with the homonymous loss of Eq.~\ref{eq:compr_entropy}--\ref{eq:compr_repr}) measures the difference between the predicted probabilities for the gold (positive) labels obtained when the model is fed with the full text ($\widehat{Y_+}^f$) 
and when it is fed with the complement of the predicted rationales ($\widehat{Y_+}^c$).
We also compare classification performance (again using \emph{micro-F1}) in both cases, i.e., when considering \emph{masked inputs} (using $Z$) and \emph{complementary inputs} (using $Z^c$).\vspace{1.5mm}

\noindent\textbf{Rationale quality:} Faithful explanations (of system reasoning) are not always appropriate for users \cite{jacovi2020}, thus we also evaluate rationale quality from a user perspective. The latter can be performed in two ways. \emph{Objective} evaluation compares predicted rationales with gold annotations, typically via \emph{Recall, Precision, F1} (comparing system-selected to human-selected facts in our case). In \emph{subjective} evaluation, human annotators review the extracted rationales. We opt for an objective evaluation, mainly due to lack of resources. As rationale sparsity (number of selected paragraphs) differs across methods, which affects \emph{Recall, Precision, F1}, we evaluate rationale quality with \emph{mean R-Precision} ({\small mRP}) \cite{Manning2009}. That is, for each case, the model ranks the paragraphs it selects by decreasing confidence, and we compute Precision@$k$, where $k$ is the number of paragraphs in the gold rationale; we then average over test cases. For completeness, we also report F1 (comparing predicted and gold rationale paragraphs), although it is less fair, because of the different sparsity of different methods, as noted.

\begin{table}[h]
\footnotesize{
    \centering
    \resizebox{\columnwidth}{!}{
    \begin{tabular}{p{4.2cm}|C{1.2cm}|c}
         \multirow{2}{*}{\echr article} & Training & Classification \\
         &  cases &  F1 $\uparrow$ \\ \hline
         2 - Right to life & 623 & 78.3 $\pm$ 2.3 \\
         3 - Prohibition of torture & 1740 & 85.9 $\pm$ 0.9 \\
         5 - Right to liberty and security & 1623 & 81.1 $\pm$ 1.5 \\
         6 - Right to a fair trial & 5437 & 80.1 $\pm$ 1.0 \\
         8 - Right to respect for private 
        %  \& family 
        life & 1056 & 72.5 $\pm$ 1.8 \\
         10 -  Freedom of expression & 441 & 77.4 $\pm$ 1.6 \\
         11 -  Freedom of assembly 
        %  and association 
        & 162 & 72.1 $\pm$ 3.3 \\
         13 - Right to an effective remedy & 1665 & 29.2 $\pm$ 3.3 \\
         14 - Prohibition of discrimination & 444 & 44.8 $\pm$ 7.4 \\
         34 - Individual applications & 547 & 10.0 $\pm$ 5.0 \\
         46 - Binding force 
        %  \& execution 
        of judgments & 187 & 2.6 $\pm$ 3.2 \\
         P1-1 - Protection of property & 1558 & 77.9 $\pm$ 1.3 \\ \hline
         Rest of the articles & $<$ 100 & $<$ 50.0 \\ \hline
         \multicolumn{1}{l}{Overall performance (micro-F1)} &  & 72.7 $\pm$ 1.2 \\
    \end{tabular}
    }
    \vspace*{-2mm}
    \caption{Classification performance of \hierbertmax (no mask) across \echr articles on development data, with respect to the number of training cases (instances).} 
    \vspace*{-4mm}
    \label{tab:class}
    }
\end{table}

\begin{table*}[t!]
    \centering
    \resizebox{\textwidth}{!}{
    \begin{tabular}{l|c|c|c|c|c|c}
        \multirow{2}{*}{Method}  & sparsity & Entire Input &  \multicolumn{2}{c|}{Masked Input ($Z$)}  & \multicolumn{2}{c}{Compl.\ Input ($Z^c$)} \\
       &  (aim: 30\%) &   micro-F1 $\uparrow$ & micro-F1 $\uparrow$ & Suff. $\downarrow$ & micro-F1 $\downarrow$ & Comp. $\uparrow$ \\ \hline
        \textsc{random classifier}                               & -               & 30.8 $\pm$ 0.3 & \multicolumn{4}{c}{-} \\
        \hierbertmax (no masking)                                                                & -               & \textbf{73.7 $\pm$ 0.6} & \multicolumn{4}{c}{-} \\ \hline
         \textsc{hierbert-ha} + $L_s$ (Eq.~\ref{eq:sparsity})                           \cite{lei2016}    & 31.7 $\pm$ 1.1  & 73.1 $\pm$ 0.6 & 69.5 $\pm$ 2.4 & 0.063 & 58.8 $\pm$ 1.5 & 0.181 \\
         \textsc{hierbert-ha} + $L_s$ + $L_g$ (Eq.~\ref{eq:compr_entropy})              \cite{yu2019}     & 31.4 $\pm$ 1.9  & 72.8 $\pm$ 0.6 & 68.1 $\pm$ 4.4 & 0.069 & 59.0 $\pm$ 1.5 & 0.171 \\
         \textsc{hierbert-ha} + $L_s$ + $L_g$ (Eq.~\ref{eq:compr_repr})                 (ours)            & 31.4 $\pm$ 1.3  & 72.6 $\pm$ 1.5 & 69.8 $\pm$ 0.8 & 0.043 & 59.6 $\pm$ 2.7  & 0.156 \\
         \textsc{hierbert-ha} + $L_s$ + $L_r$ (Eq.~\ref{eq:compr_prob},~\ref{eq:sing})  (ours)            & 31.5 $\pm$ 0.8  & 72.8 $\pm$ 0.5 & \textbf{70.5 $\pm$ 0.8} & \textbf{0.040} & \textbf{55.9 $\pm$ 2.8} & \textbf{0.204} \\ \hline
         \textsc{hierbert-ha} + rationale supervizion (Eq.~\ref{eq:noisy})                                & 33.1 $\pm$ 6.0  & 73.1 $\pm$ 0.5 & 69.2 $\pm$ 1.1 & 0.053 & 56.7 $\pm$ 6.6 & 0.191 \\
    \end{tabular}
    }
    \caption{\emph{Classification performance} (classification micro-F1) and \emph{faithfulness} results on test data. Faithfulness is measured by considering \emph{Sufficiency} (Suff.) and \emph{Comprehensiveness} (Comp.), i.e., how close the label probabilities of the model are when using the rationales (masked input) or the complements of the rationales (complementary input), respectively, as opposed to using the entire input. Lower Suff.\ ($\downarrow$) and higher Comp.\ ($\uparrow$) are better. We also report micro-F1 for the masked and complementary input; higher and lower F1, respectively, are better.}
    \label{tab:test}
    \vspace*{-4mm}
\end{table*}

\section{Experimental Results}

\subsection{Initial Classification Performance}

Table~\ref{tab:class} reports the classification performance of \hierbertmax (no masking, no rationales), across \echr articles. F1 is 72.5\% or greater for most of the articles with 1,000 or more training instances. The scores are higher for articles 2, 3, 5, 6, because (according to the legal expert who provided the gold allegation rationales), (i) there is a sufficient number of cases regarding these articles, and (ii) the interpretation and application of these articles is more fact-dependent than those of other articles, such as articles 10 or 11 \cite{oboyle2018}. On the other hand, although there is a fair amount of training instances for articles 13, 14, 34, and 46, these articles are triggered in a variety of ways, many of which turn on legal procedural technicalities.

\subsection{Tuning the $\lambda$ Hyper-parameters}

Instead of tuning simultaneously all the  $\lambda_*$ hyper-parameters of Eq.~\ref{eq:totalLoss}, we adopt a greedy, but more intuitive strategy: we tune one $\lambda$ at a time, fix its value, and proceed to the next; $\lambda${s} that have not been tuned are set to zero, i.e., the corresponding regularizer is not used yet. We begin by tuning $\lambda_s$, aiming to achieve a desirable level of sparsity without harming classification performance. We set the sparsity threshold of $L_s$ (Eq.~\ref{eq:sparsity}) to $T= 0.3$ (select approx.\ 30\% of the facts), which is the average sparsity of the silver rationales (Table~\ref{tab:dataset}).  We found  $\lambda_s = 0.1$ achieves the best overall results on development data, thus we use this value for the rest of the experiments.\footnote{Consult Appendix~\ref{sec:appendix_d} for more detailed results.} 
To check our hypothesis that continuity ($L_c$) is not beneficial in our task, we tuned $\lambda_c$ on development data, confirming that the best overall results are obtained for $\lambda_c = 0$.\footnotemark[9] Thus we omit $L_c$ in the rest of the experiments.

\subsection{Comprehensiveness/Singularity Variants}

Next, we tuned and compared the variants of the comprehensiveness loss $L_g$ (Table~\ref{tab:comprehensiveness}). Targeting the label probabilities (Eq.~\ref{eq:compr_prob}) instead of the losses (Eq.~\ref{eq:compr_entropy}) leads to lower rationale quality. Targeting the document representations (Eq.~\ref{eq:compr_repr}) has the best rationale quality results, retaining  (as with all versions of $L_g$) the original classification performance (micro-F1) of Table~\ref{tab:class}. Hence, we keep the $L_g$ variant of Eq.~\ref{eq:compr_repr} in the remaining experiments of this section, with the corresponding $\lambda_g$ value ($1\mathrm{e}\text{-}3$).  

\begin{table}[h]
\footnotesize{
    \centering
    \resizebox{\columnwidth}{!}{
    \begin{tabular}{c|c|c|c|c}
        $L_g$ & classification & sparsity &  \multicolumn{2}{c}{rationale quality} \\
        variant &  micro-F1 $\uparrow$ & (aim: 30\%) & F1 $\uparrow$ & mRP $\uparrow$ \\ \hline
        Eq.~\ref{eq:compr_entropy} & 73.0 $\pm$ 0.5 & 31.4 $\pm$ 1.9  & 35.4 $\pm$ 5.8 & 38.4 $\pm$ 5.9 \\
        Eq.~\ref{eq:compr_prob} & \textbf{73.1 $\pm$ 0.7}  & 31.9 $\pm$ 1.4 & 30.3 $\pm$ 3.0 & 32.6 $\pm$ 2.6 \\
        Eq.~\ref{eq:compr_repr} & 72.8 $\pm$ 0.8 & 31.8 $\pm$ 1.3 & \textbf{38.3 $\pm$ 2.3} & \textbf{41.2 $\pm$ 2.1} \\
    \end{tabular}
    }
    \vspace*{-2mm}
    \caption{Development results for variants of $L_g$ (\emph{comprehensiveness}) and varying $\lambda_g$ values (omitted).
    }
    \label{tab:comprehensiveness}
    \vspace*{-5mm}
    }
\end{table}

\begin{table}[h]
\footnotesize{
    \centering
    \resizebox{\columnwidth}{!}{
    \begin{tabular}{c|c|c|c|c}
        $L_r$ & classification & sparsity &  \multicolumn{2}{c}{rationale quality} \\
        variant &  micro-F1  $\uparrow$ & (aim: 30\%) & F1 $\uparrow$ & mRP $\uparrow$ \\ \hline
        Eq.~\ref{eq:compr_entropy},~\ref{eq:sing} & \textbf{73.4 $\pm$ 0.8} & 32.8 $\pm$ 2.8 & 36.9 $\pm$ 3.6 & 39.0 $\pm$ 3.9 \\
        Eq.~\ref{eq:compr_prob},~\ref{eq:sing} & 72.5 $\pm$ 0.7  & 32.0 $\pm$ 1.0 & \textbf{39.7 $\pm$ 3.1} & \textbf{42.6 $\pm$ 3.8} \\
        Eq.~\ref{eq:compr_repr},~\ref{eq:sing} & 72.8 $\pm$ 0.3 & 31.5 $\pm$ 0.9 & 33.0 $\pm$ 2.7 & 35.5 $\pm$ 2.6 \\
    \end{tabular}
    }
    \vspace*{-2mm}
    \caption{Development results for variants of $L_c$ (\emph{singularity}) and varying $\lambda_r$ values (omitted).}
    \vspace*{-1mm}
    \label{tab:singularity}
    }
\end{table}

Concerning the singularity loss $L_r$ (Table~\ref{tab:singularity}), targeting the label probabilities (Eq.~\ref{eq:compr_prob},~\ref{eq:sing}) provides the best rationale quality, comparing to all the methods considered. Interestingly Eq.~\ref{eq:compr_repr}, which performed best in $L_g$ (Table~\ref{tab:comprehensiveness}), does not perform well in $L_r$, which uses $L_g$ (Eq.~\ref{eq:sing}). We suspect that in $L_r$, where we use a random mask $Z^r$ that may overlap with $Z$, requiring the two document representations $D_M, D_M^r$ to be dissimilar (when using Eq.~\ref{eq:compr_repr},~\ref{eq:sing}) may be a harsh regularizer with negative effects.

\subsection{Task Performance and Faithfulness}

Table~\ref{tab:test} presents results on test data. The models that use the hard attention mechanism and are regularized to extract rationales under certain constraints (\hierbertha + $L_*$) have comparable classification performance to \hierbertmax. Furthermore, although paragraph embeddings are contextualized and probably have some information leak for all methods, our proposed extensions in rationale constraints better approximate faithfulness, while also respecting sparsity. Our proposed extensions lead to low sufficiency (lower is better, $\downarrow$), i.e., there is only a slight deterioration in label probabilities when we use the predicted rationale instead of the whole input. They also lead to high comprehensiveness (higher is better, $\uparrow$); we see a 20\% deterioration in label probabilities when using the complement of the rationale instead of the whole input. Interestingly, our variant with the singularity loss (Eq.~\ref{eq:compr_prob}, \ref{eq:sing}) is more faithful than the model that uses supervision on silver rationales (Eq.~\ref{eq:noisy}).

\begin{table}[h]
    \centering
    \resizebox{\columnwidth}{!}{
    \begin{tabular}{l|c|c|c|c}
        \multirow{2}{*}{Method} & \multicolumn{2}{c|}{Silver rationales (31\%)} & \multicolumn{2}{c}{Gold rationales (36\%)} \\
        &  mRP $\uparrow$ &  F1 $\uparrow$ & mRP $\uparrow$ &  F1 $\uparrow$\\  \hline
        \textsc{random rationale}                                           & 30.2 $\pm$ 1.1 & 27.8 $\pm$ 1.1 & 35.1 $\pm$ 1.7   & 30.2 $\pm$ 2.2 \\ \hline
        \multicolumn{5}{c}{\hierbertha} \\ \hline
        + $L_s$ (Eq.~\ref{eq:sparsity})                           & 43.1 $\pm$ 6.5 & 37.3 $\pm$ 5.4 & 51.9 $\pm$ 5.7   & 45.7 $\pm$ 5.4 \\
        + $L_s$ + $L_g$ (Eq.~\ref{eq:compr_entropy})              & 41.0 $\pm$ 5.1 & 37.5 $\pm$ 6.7 & 48.9 $\pm$ 6.5   & 44.5 $\pm$ 6.8 \\
        + $L_s$ + $L_g$ (Eq.~\ref{eq:compr_repr})                 & 43.0 $\pm$ 1.5 & 38.5 $\pm$ 1.9 & 50.9 $\pm$ 3.2   & 45.8 $\pm$ 3.3 \\
        + $L_s$ + $L_r$ (Eq.~\ref{eq:compr_prob},~\ref{eq:sing})  & \textbf{45.1 $\pm$ 2.1} & \textbf{40.9 $\pm$ 2.5} & \textbf{53.6 $\pm$ 2.3}   & \textbf{48.3 $\pm$ 1.2} \\  \hline
        + supervision  (Eq.~\ref{eq:noisy})             & 43.1 $\pm$ 5.0 & 39.1 $\pm$ 7.1 & 51.4 $\pm$ 6.7   & 46.8 $\pm$ 0.5 \\ 
    \end{tabular}
    }
        \vspace*{-2mm}
    \caption{\emph{Rationale quality} results on the
    50 test cases that have both silver and gold allegation rationales.
    Average silver/gold rationale sparsity (\%) in brackets.
    }
    \vspace*{-5mm}
    \label{tab:test2}
\end{table}

\subsection{Rationale Quality}
We now consider rationale quality, focusing on \hierbertha variants without rationale supervision. Similarly to our findings on development data (Tables \ref{tab:comprehensiveness}, \ref{tab:singularity}), we observe (Table~\ref{tab:test2}) that using (a) our version of \emph{comprehensiveness} loss (Eq.~\ref{eq:compr_repr}) or (b) our \emph{singularity} loss (Eq.~\ref{eq:compr_prob},~\ref{eq:sing}) achieves better results compared to former methods, and (b) has the best results. The \emph{singularity} loss is better in both settings (silver or gold test rationales), even compared to a model that uses supervision on silver rationales. The random masking of the singularity loss, which guides the model to learn to extract masks that perform better than \emph{any} other mask, proved to be particularly beneficial in rationale quality. Similar observations are derived given the results on the full test set considering silver rationales.\footnote{See Appendix~\ref{sec:appendix_d} for rationale quality evaluation on the full test set.} In general, however, we observe that the rationales extracted by all models are far from human rationals, as indicated by the poor results ({\small mRP, F1}) on both silver and gold rationales. Hence, there is ample scope for further research.

\subsection{Qualitative Analysis}

\noindent\textbf{Quality of silver rationales:} Comparing silver rationales with gold ones, annotated by the legal expert, we find that silver rationales are not complete, i.e., they are usually fewer than the gold ones. They also include additional facts that have not been annotated by the expert. According the expert, these facts do not support allegations, but are included for technical reasons (e.g., \emph{``The national court did not accept the applicant's allegations.''}). Nonetheless, ranking methods by their rationale quality measured on silver rationales produces the same ranking as when measuring on gold rationales in the common subset of cases (Table~\ref{tab:test2}). Hence, it may be possible to use silver rationales, which are available for the full dataset, to rank systems participating in \ecthr rationale generation challenges.\vspace{2mm} 

\noindent\textbf{Model bias:} Low {\small mRP} with respect to gold rationales means that the models rely partially on non causal reasoning, i.e., they select secondary facts that do not justify allegations according to the legal expert. In other words, the models are sensitive to specific language, e.g., they misuse (are easily fooled by) references to health issues and medical examinations as support for Article 3 alleged violations, or references to appeals in higher courts as support for Article 5, even when there is no concrete evidence.\footnote{See Appendix~\ref{sec:appendix_f} for examples of \ecthr cases.} Manually inspecting the predicted rationales, we did not identify bias on demographics. Although such spurious features may be buried in the contextualized paragraph encodings ($P^{\cls}_i$).
In general, \emph{de-biasing} models could benefit rationale extraction and we aim to investigate this direction in future work \cite{huang2020}.\vspace{2mm}

\noindent\textbf{Plausibility:} \emph{Plausibility} refers to how convincing the interpretation is to humans \cite{jacovi2020}. While the legal expert annotated all relevant facts with respect to allegations, according to his manual review, allegations can also be justified by sub-selections (parts) of rationales. Thus, although a method may fail to extract all the available rationales, the provided (incomplete) set of rationales may still be a convincing explanation. To properly estimate plausibility across methods, one has to perform a subjective human evaluation which we did not conduct due to lack of resources.

\section{Conclusions and Future work}

We introduced a new application of rationale extraction in a new legal text classification task concerning alleged violations on \ecthr cases. We also released a dataset for this task to foster further research. Moreover, we compared various rationale constraints in the form of regularizers and introduced a new one (\emph{singularity}) improving faithfulness and rationale quality in a paragraph-level setup comparing both with silver and gold rationales.

In the future, we plan to investigate more constraints that may better fit paragraph-level rationale extraction and explore techniques to de-bias models and improve rationale quality. Paragraph-level rationale extraction can be also conceived as self-supervised extractive summarization to denoise long documents, a direction we plan to explore in the challenging task of case law retrieval~\cite{locke2018}. 

%%%%%%%%%%%%%%%%%%%%%%%%%%%%%%%%%%%%%%%%%%%%%%%%%%%%
\section*{Acknowledgments}
We would like to thank the anonymous reviewers (esp. reviewer \#2) for their constructive detailed comments. Nikolaos Aletras is supported by EPSRC grant EP/V055712/1, part of the European Commission CHIST-ERA programme, call 2019 XAI: Explainable Machine Learning-based Artificial Intelligence.

%%%%%%%%%%
\bibliography{naacl2021}
\bibliographystyle{acl_natbib}

\appendix
\section{Why is the new dataset any different?}
\label{sec:appendix_a}
\noindent The new dataset is noteworthy for three reasons:

\smallskip
\noindent\textbf{Legal rationale dataset:} 
This is the first rationale extraction dataset for the legal domain, where justifying decisions is essential and often requires complicated reasoning. Thus the dataset is a challenging test-bed that will boost rationale extraction research. Predicting and justifying alleged article violations is also helpful in practice and would assist legal judgment prediction \cite{Aletras2016}. 

\smallskip
\noindent\textbf{Paragraph-level rationales:}
Each case description is a carefully planned  document of multiple paragraphs enumerating facts chronologically. Each paragraph concisely provides information at a granularity considered appropriate for legal reasoning. Accordingly, rationales must be extracted at this granularity, either selecting an entire factual paragraph or not, as opposed to most rationale extraction datasets where particular words or phrases can be selected (e.g., from product reviews).

% \smallskip 
% \noindent\textbf{Judgment prediction supported:} The dataset can also be used for legal judgment prediction, since it also contains the rulings on the alleged article violations, as already noted. Hence, it can also connect work on allegation prediction with judgment prediction, for example by developing supportive tools that would allow plaintiffs to select their allegations and predict likely judgments for them. 

\section{Baseline Model}
\label{sec:appendix_b}

In preliminary experiments, we found that the proposed model (41.9M parameters) (Table~\ref{tab:encoder}, third row) relying on a shared paragraph encoder (\hierbert) to produce both context-aware representations and rationales, has comparable classification performance and better rationale quality compared to: (i) a model with two independent paragraph encoders  (82.8M parameters) (Table~\ref{tab:encoder}, first row), similar to the one used in the literature \cite{lei2016, yu2019, jain2020}; (ii) a model that omits the $Q$ and $K$ projection layers  (41.4M parameters) (Table~\ref{tab:encoder}, second row). Recall that \citet{lei2016}, \citet{yu2019}, \citet{jain2020} extract rationales at the word-level, and their encoders, either \textsc{bilstm}{s} \citep{lei2016, yu2019} or \textsc{bert} \citep{jain2020}, operate on that level of granularity. 

\begin{table}[h]
 \footnotesize{
    \centering
    \resizebox{\columnwidth}{!}{
    \begin{tabular}{l|c|c|c|}
        \multirow{2}{*}{Method} & classification & \multicolumn{2}{c|}{Silver rationales} \\
          &  micro-F1 & mRP &  F1 \\  \hline
        2x \hierbert                                           & 73.4 $\pm$ 0.6 & 35.1 $\pm$ 7.9 & 29.3 $\pm$ 8.7  \\
        1x \hierbert excl. ($Q$,$K$)                           & \textbf{73.5 $\pm$ 0.7} & 29.2 $\pm$ 7.9 & 26.4 $\pm$ 7.9 \\
        1x \hierbert + ($Q$,$K$)                               & 73.2 $\pm$ 0.5 & \textbf{35.9 $\pm$ 4.7} & \textbf{39.0 $\pm$ 4.9} \\
    \end{tabular}
    }
    \caption{Results on classification performance and rationale quality on development data.}
    \label{tab:encoder}
    }
\end{table} 

\section{Annotation of Gold Rationales}
\label{sec:appendix_c}

The full dataset has the following characteristics:

\begin{itemize}
    \setlength\itemsep{0.5mm}
    \item There are 1,000 cases in the test set. These are the most recent and have been ruled from October 5, 2017 until July 7, 2019.
    \item The average number of facts (paragraphs) per case is 25.2 ranging from 5 to 259.
    \item Almost half of the cases concern applications against 6 European states (defendants): Russia (229), Turkey (122), Ukraine (80), Romania (47), Moldova (44), Lithuania (43). Number of test cases in brackets.
    \item The allegations in the vast majority of cases (approx. 88\%) concern nine articles: `6' (394), `3' (233), `5' (197), `8' (188), `P1-1' (155), `13' (123), `35' (107), `10' (106), `2' (76).  Number of test cases in brackets.
\end{itemize}

Based on the above statistics and the opinion of the legal expert, for the gold rationales we considered a subset of 50 cases with the following characteristics: 
\begin{itemize}
    \setlength\itemsep{0.5mm}
    \item Each case should consist of 25 ± 10 facts.
    \item The cases should be as representative as possible with respect to the defendants (European states).
    \item The cases should have allegations in a subset of the following articles \{2, 3, 5, 6\}, whose interpretation is more fact-dependent based on the literature \cite{oboyle2018} and our presented empirical results (Table~\ref{tab:class}).
\end{itemize}

The annotation guidelines briefly were: 
\begin{itemize}
    \setlength\itemsep{0.5mm} 
    \item The annotator (legal expert) inspects (reads) all the factual paragraphs of the case and selects one or more articles in the predefined set \{2, 3, 5, 6\}, that should have been argued by the applicants according to the text.
    \item The annotator selects the factual paragraphs that ``clearly'' indicate allegations for the selected article(s), annotated in the first step.
\end{itemize}

The legal expert performance compared to the gold allegedly violated articles, is 92.3\% micro-F1 (Table~\ref{tab:ann_cases}). In few cases, the legal expert selected more articles (hypothesized allegations for articles 3 and 5) compared to the gold ones. As he suggested, it is a common trend for the applicants, based on the legal opinion of their attorneys, to raise allegations only for a few articles that they believe can be justified and proved to be violated, i.e, if a citizen has no concrete evidence (documents) for his torture, his lawyer may suggest him to not raise this issue in his application. The legal expert also missed a few allegations for articles 2 and 6. The best of our models, (\hierbertsa + $L_s$ + $L_r$) achieves 87.6\% micro-F1 in the same subset.

\begin{table}[h]
    \centering
    \resizebox{\columnwidth}{!}{
    \begin{tabular}{l|c|c}
         \echr article & Expert F1 & Model F1  \\
         \hline
         2 - Right to life & 88.9 & 82.4 \\
         3 - Prohibition of torture & 95.5 & 92.7 \\
         5 - Right to liberty and security & 85.7 & 88.0 \\
         6 - Right to a fair trial & 95.0 & 84.2 \\
         \hline
         micro-F1 & 92.3 & 87.6 \\ 
         macro-F1 & 91.3 & 86.8 \\
    \end{tabular}
    }
    \caption{Classification performance of the legal expert and our best method on the 50 annotated test cases, considering only the facts of each case.}
    \label{tab:ann_cases}
\end{table}

\section{Additional Experimental Results}
\label{sec:appendix_d}
For completeness, we report results on development data for \emph{sparsity} loss ($L_s$) in Table~\ref{tab:sparsity} and \emph{continuity} loss ($L_c$) in Table~\ref{tab:continuity} for different values of  $\lambda_*$ hyper-parameters.

\begin{table}[h]
\footnotesize{
    \centering
    \resizebox{\columnwidth}{!}{
    \begin{tabular}{c|c|c|c|c}
        & classification & sparsity &  \multicolumn{2}{c}{rationale quality} \\
        $\lambda_s$  &  micro-F1 $\uparrow$ & (aim: 30\%) & F1 $\uparrow$ & mRP $\uparrow$ \\ \hline
        0 & \textbf{73.3 $\pm$ 0.9} & 90.3 $\pm$ 19.3 & 32.4 $\pm$ 4.0 & 35.1 $\pm$ 5.5 \\
        0.01 & 72.9 $\pm$ 0.4 & 30.3 $\pm$ 2.7 & 36.6 $\pm$ 10.8 & 36.8 $\pm$ 8.8 \\
        0.1 & 73.2 $\pm$ 0.5 & 31.7 $\pm$ 1.1 & \textbf{39.1 $\pm$ 4.7} & \textbf{39.0 $\pm$ 4.9} \\
        0.5 & 71.3 $\pm$ 0.9 & 37.6 $\pm$ 7.5 & 29.1 $\pm$ 8.2 & 30.0 $\pm$ 8.4 \\
        1.0 & 71.1 $\pm$ 1.7 & 35.7 $\pm$ 5.6 & 36.2 $\pm$ 8.1 & 38.7 $\pm$ 7.6 \\
    \end{tabular}
    }
    \caption{Development results varying $\lambda_s$ (\emph{sparsity}).} 
    \label{tab:sparsity}
    }
\end{table}

\begin{table}[h]
\footnotesize{
    \centering
    \resizebox{\columnwidth}{!}{
    \begin{tabular}{c|c|c|c|c}
        & classification & sparsity &  \multicolumn{2}{c}{rationale quality} \\
        $\lambda_c$  &  micro-F1 $\uparrow$ & (aim: 30\%) & F1 $\uparrow$ & mRP $\uparrow$ \\ \hline
        0 & \textbf{73.2 $\pm$ 0.5} & 31.7 $\pm$ 1.1 & \textbf{35.9 $\pm$ 4.7} & \textbf{39.0 $\pm$ 4.9} \\
        0.01 & \textbf{73.2 $\pm$ 0.8} & 31.3 $\pm$ 2.5 & 30.9 $\pm$ 6.9 & 34.1 $\pm$ 6.3 \\
        0.1 & 72.8 $\pm$ 0.5 & 49.4 $\pm$ 20.6 & 26.1 $\pm$ 8.2 & 23.9 $\pm$ 1.6 \\
    \end{tabular}
    }
    \caption{Development results varying $\lambda_c$ (\emph{continuity}). 
    }
    \label{tab:continuity}
    }
    \vspace{-4mm}
\end{table}

In Section 5.6, we reported rationale quality on a subset of test data that includes silver and gold allegation rationales. For completeness, in Table~\ref{tab:test3} we report results on the full set of test data for silver rationales. We observe that all findings and particularly the ranking of the methods with respect to the subset of silver and gold rationales hold. Furthermore, we observe that the rationale quality performance on the full test set is slightly inferior in most cases (2-4\%), which is expected as the sample annotated by the expert included only cases with allegations for articles that are more explainable.

\begin{table}[h]
 \footnotesize{
    \centering
    \resizebox{\columnwidth}{!}{
    \begin{tabular}{l|c|c|c}
        \multirow{2}{*}{Method} & \multicolumn{2}{c|}{Silver rationales (31\%)} \\
          &  mRP $\uparrow$ &  F1 $\uparrow$ \\  \hline
        \textsc{random}                                           & 30.7 $\pm$ 0.7 & 26.2 $\pm$ 0.5  \\ \hline
        \hierbertha + $L_s$ (Eq. 1)                           & 39.0 $\pm$ 3.9 & 35.1 $\pm$ 3.7 \\
        \hierbertha + $L_s$ + $L_g$ (Eq. 3)              & 39.1 $\pm$ 5.6 & 34.7 $\pm$ 5.7 \\
        \hierbertha + $L_s$ + $L_g$ (Eq. 5)                 & 42.7 $\pm$ 1.8 & 38.2 $\pm$ 1.5  \\
        \hierbertha + $L_s$ + $L_r$ (Eq. 6)  & \textbf{43.3 $\pm$ 2.3} & \textbf{39.0 $\pm$ 2.1} \\
    \end{tabular}
    }
    \caption{Results on rationale quality on the full set of  test data for silver rationales.}
    \label{tab:test3}
    \vspace{-4mm}
    }
\end{table} 

\section{Using ECtHR dataset via \huggingface}
\label{sec:appendix_e}

The dataset is available at \url{https://archive.org/details/ECtHR-NAACL2021}; but you can easily load and use it in python with two lines of code:
\footnotesize{
\begin{verbatim}

from datasets import load_dataset 
dataset = load_dataset("ecthr_cases")
\end{verbatim}
}

\section{Examples of extracted Rationales from \ecthr cases with comments}
\label{sec:appendix_f}

\begin{figure*}[h]
\begin{center}
\includegraphics[width=\textwidth]{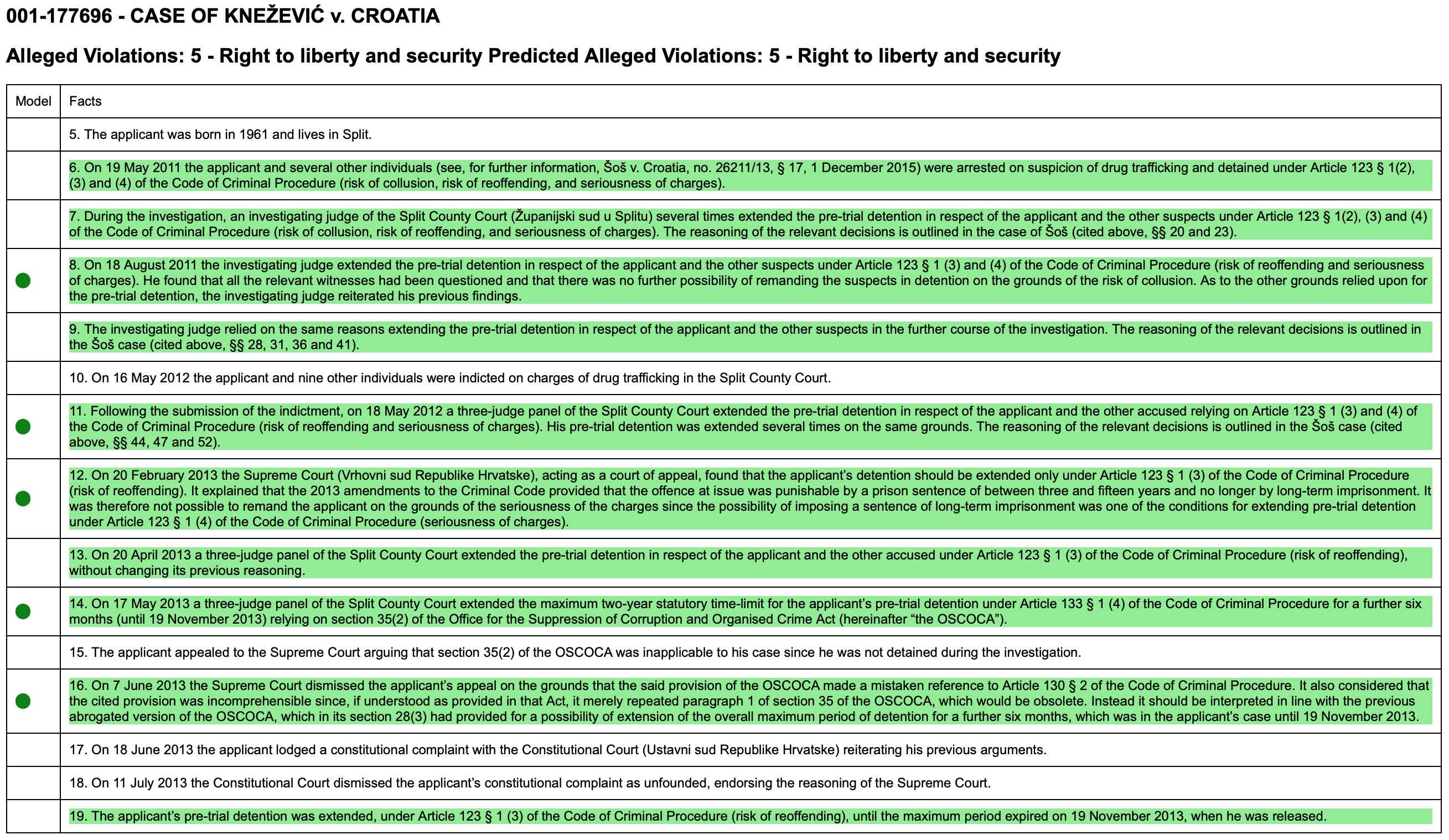}
\caption{(\href{https://hudoc.echr.coe.int/eng\#\%7B"itemid":["001-177696"]\%7D}{KNEZEVIC v. CROATIA, No. 55133/13\})} The model extracted most of the relevant facts indicating a possible violation of Article 5. Note that 67\% (10 of 15) of the facts were considered relevant by the legal expert. Our model has a disadvantage in this case because, being trained to operate at a predefined sparsity level (30\%), it extracted only 5 of the 15 facts (33\%).}
\label{fig:case3}
\end{center}
\end{figure*}

\begin{figure*}[h]
\begin{center}
\includegraphics[width=\textwidth]{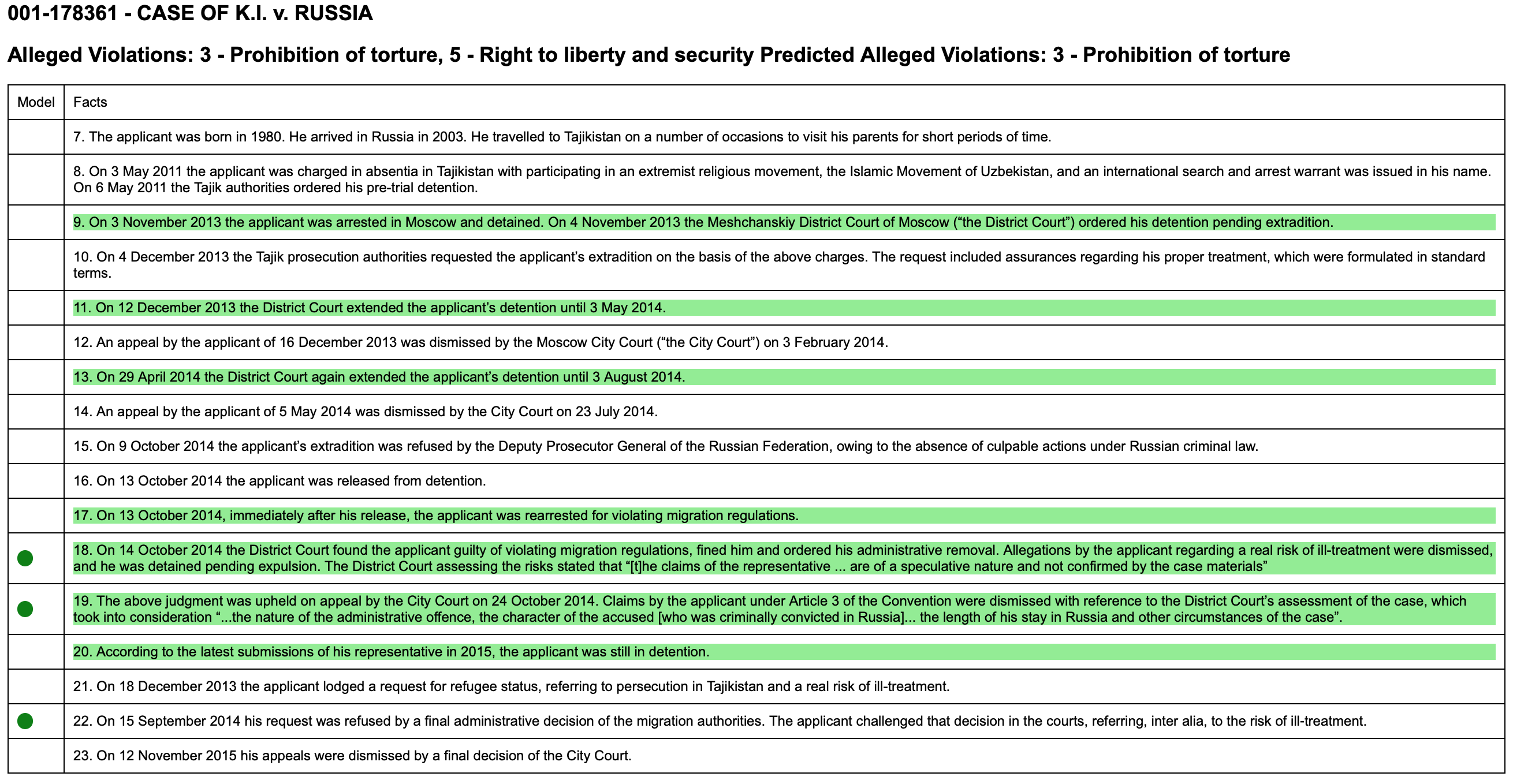}
\end{center}
\caption{(\href{https://hudoc.echr.coe.int/eng\#\%7B"itemid":["001-178361"]\%7D}{K.I. v. RUSSIA, No. 58182/14}) Paragraphs 9, 11, 13 and 20 clearly indicate plausible violation of the right to liberty (Article 5), as they refer to continuous extension of applicant detention, but our model was unable to extract them, thus it was unable to predict this allegation. The model targeted only paragraphs that indicate ill-treatment, which is connected to plausible violation of Article 3 (Prohibition of Torture).}
\label{fig:case4}
\end{figure*}

\begin{figure*}[h]
\begin{center}
\includegraphics[width=\textwidth]{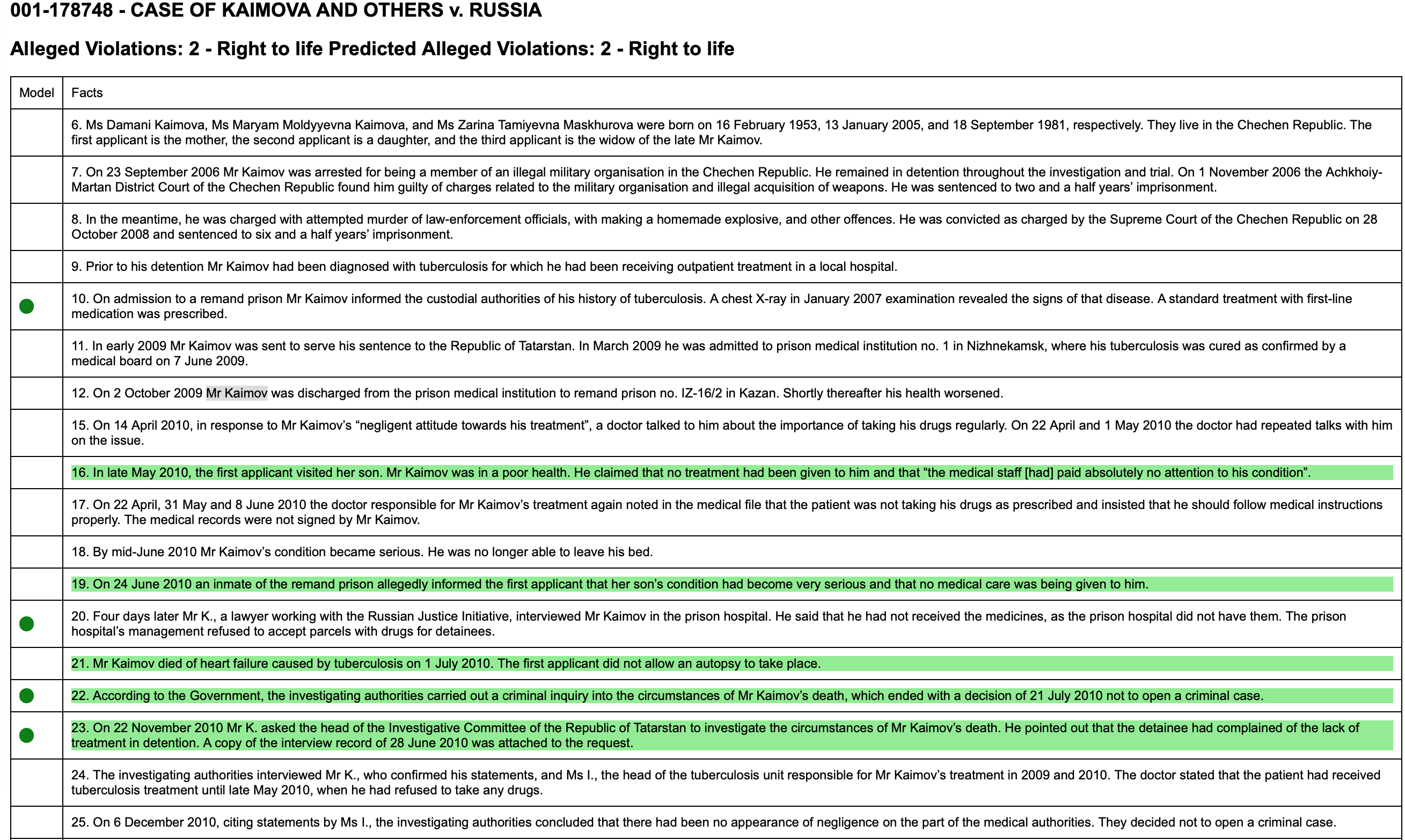}
\end{center}
\caption{(\href{https://hudoc.echr.coe.int/eng\#\%7B"itemid":["001-178748"]\%7D}{KAIMOVA AND OTHERS v. RUSSIA, No. 58182/14}) Paragraphs 16 and 19 clearly indicate that the applicant's health (life) was at risk and authorities did not pay attention, but these paragraphs were not selected by the model. Instead paragraph 10 states that the applicant initially informed the authorities for his medical history and they provided medication. This is an indication of model sensitivity to language describing health issues (tuberculosis) in general and not specific well-defined allegations for ill-treatment on the merits.}
\label{fig:case5}
\end{figure*}

\begin{figure*}[h]
\begin{center}
\includegraphics[width=\textwidth]{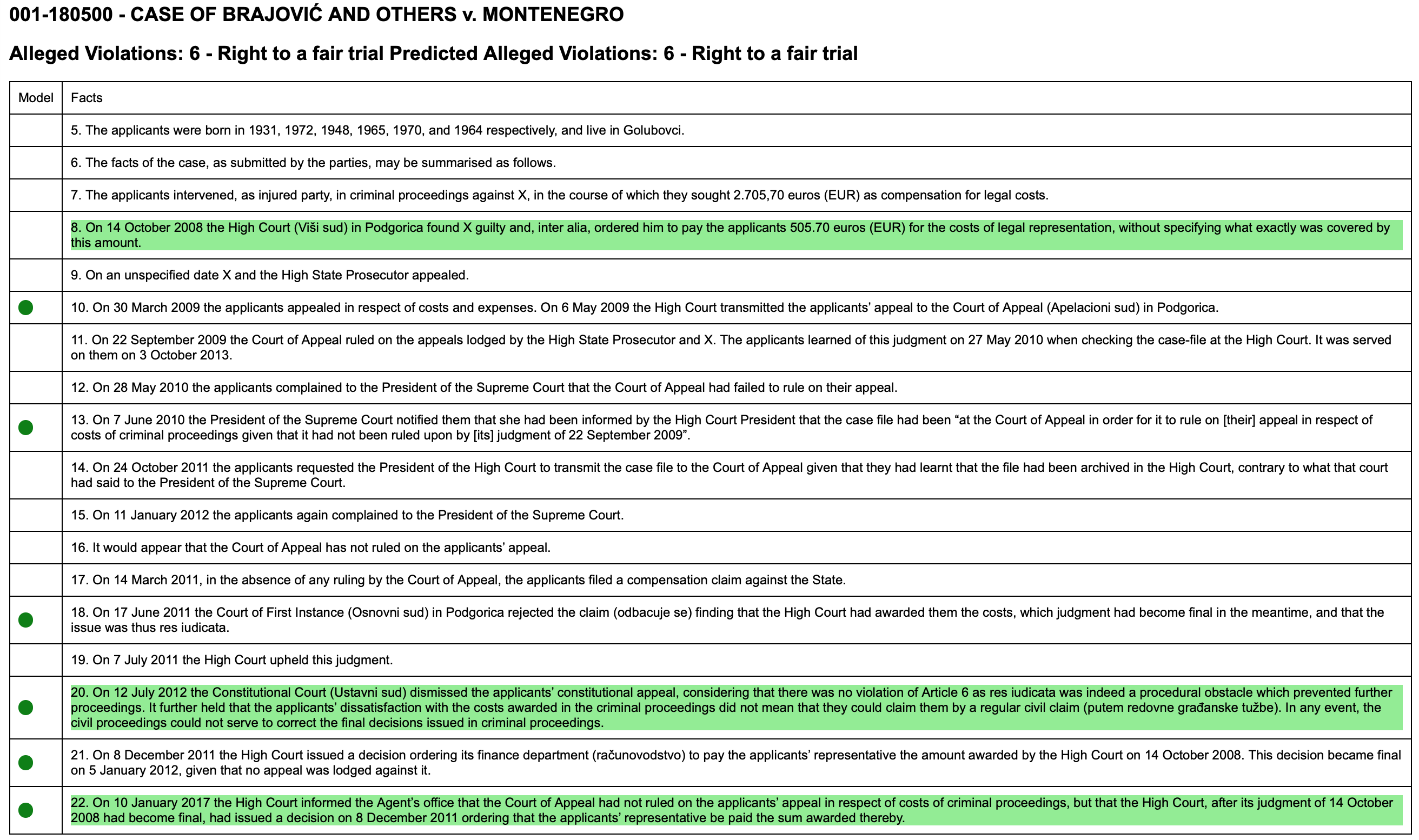}
\end{center}
\caption{(\href{https://hudoc.echr.coe.int/eng\#\%7B"itemid":["001-180500"]\%7D}{BRAJOVIC AND OTHERS v. MONTENEGRO, No. 52529/12}) A causal inference would connect paragraph 8 (initial trial) with paragraphs 20--22 (next trials) to infer mistrial, because there was is verdict after a reasonable period of time. Instead  the model seems to be sensitive to references for the involvement of higher courts as justification of mistrial (paragraphs 10, 13, 18, and 21). This suggests that the model probably follows poor (greedy) reasoning, i.e., if the applicant appealed to higher courts, then the case is mistreated.
}
\label{fig:case7}
\end{figure*}

\begin{figure*}[t]
\begin{center}
\includegraphics[width=\textwidth]{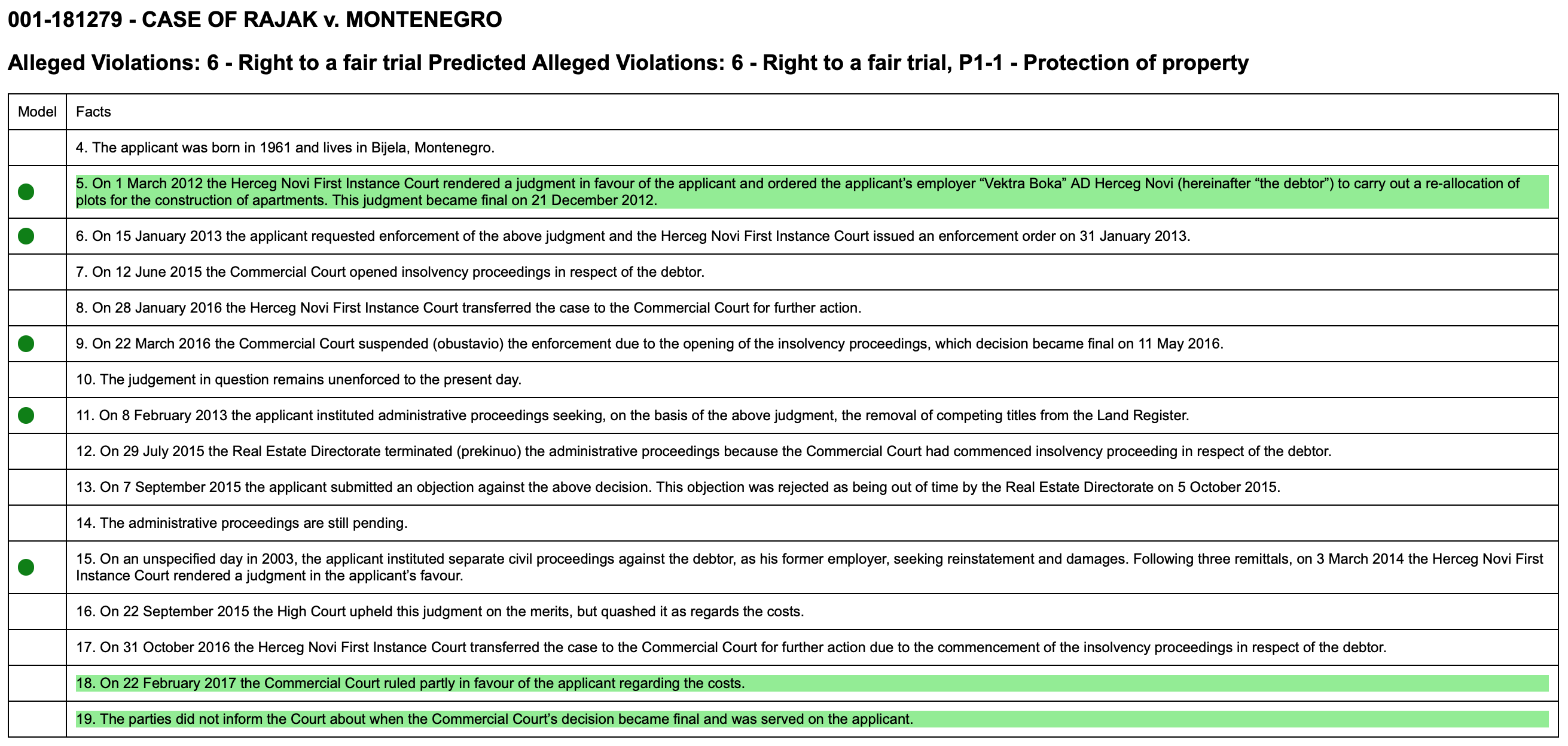}
\end{center}
\caption{(\href{https://hudoc.echr.coe.int/eng\#\%7B"itemid":["001-181279"]\%7D}{RAJAK v. MONTENEGRO, No. 71998/11}) Similarly to the case presented in Fig.~\ref{fig:case7}, the main argument in this case is mistrial because there was a verdict after a reasonable period of time (paragraphs 5 and 18-19). The model selected paragraph 11, which does not indicate plausible violations. Given the model's prediction for allegations with respect to Article 1 of the 1st Protocol on the protection of property, we believe that paragraph 11 was selected by the model as justification on that matter.}
\label{fig:case8}
\end{figure*}

\normalsize{
In Fig.~\ref{fig:case3}--\ref{fig:case8}, we present examples of \ecthr cases. The highlighting (green background colour) indicates gold rationales. The dots (green dot on the left) indicate rationales extracted by our best model, \hierbertha + $L_s$ + $L_g$ (Eq.~\ref{eq:compr_prob}, \ref{eq:sing}). In the caption of each figure, we include short comments 
% justifying on both the paragraphs that are not selected by the model (false negatives), i.e., why are these relevant?, and those that are selected but are not relevant (false positives), i.e., why are these not relevant?.
explaining false positives (paragraphs the model wrongly selected) and false negatives (paragraphs the model wrongly missed). 
}

\end{document}